\pdfoutput=1
\documentclass[11pt]{article}

\usepackage[preprint]{acl}

\usepackage{times}
\usepackage{latexsym}

\usepackage[T1]{fontenc}
\usepackage[utf8]{inputenc}

\usepackage{microtype}

\usepackage{inconsolata}

\usepackage{graphicx}
\usepackage{booktabs} % For better horizontal lines
\usepackage{array} % For table formatting
\usepackage{makecell}
\usepackage{subcaption}
\usepackage{url}
\usepackage{amsmath}
\usepackage{multirow}
\newcommand{\flan}{Flan-T5}
\newcommand{\llama}{LLaMA 2}

\title{Reasoning or a Semblance of it? \\[4pt]
A Diagnostic Study of Transitive Reasoning in LLMs}

\author{Houman Mehrafarin$^{1, 2}$\hspace{1em} Arash Eshghi$^{1}$ \hspace{1em} Ioannis Konstas$^{1}$\\
  $~^{1}$Heriot-Watt University, Edinburgh, United Kingdom \\
  $~^{2}$University of Edinburgh, Edinburgh, United Kingdom \\
  \texttt{\{hm2066, a.eshghi, i.konstas\}@hw.ac.uk}}

\begin{document}
\maketitle
\begin{abstract}

Evaluating Large Language Models (LLMs) on reasoning benchmarks demonstrates their ability to solve compositional questions. However, little is known of whether these models engage in genuine logical reasoning or simply rely on implicit cues to generate answers. In this paper, we investigate the transitive reasoning capabilities of two distinct LLM architectures, \llama~and \flan, by manipulating facts within two compositional datasets: QASC and Bamboogle. We controlled for potential cues that might influence the models’ performance, including (a) word/phrase overlaps across sections of test input; (b) models' inherent knowledge during pre-training or fine-tuning; and (c) Named Entities. Our findings reveal that while both models leverage (a), \flan~shows more resilience to experiments (b and c), having less variance than \llama. This suggests that models may develop an understanding of transitivity through fine-tuning on knowingly relevant datasets, a hypothesis we leave to future work\footnote{The code and dataset can be found at our \href{https://github.com/hmehrafarin/LLM-Reasoning-Analysis.git}{github} page.}.

\end{abstract}

\section{Introduction}

\begin{figure*}[t!]
    \centering
    \includegraphics[width=\textwidth]{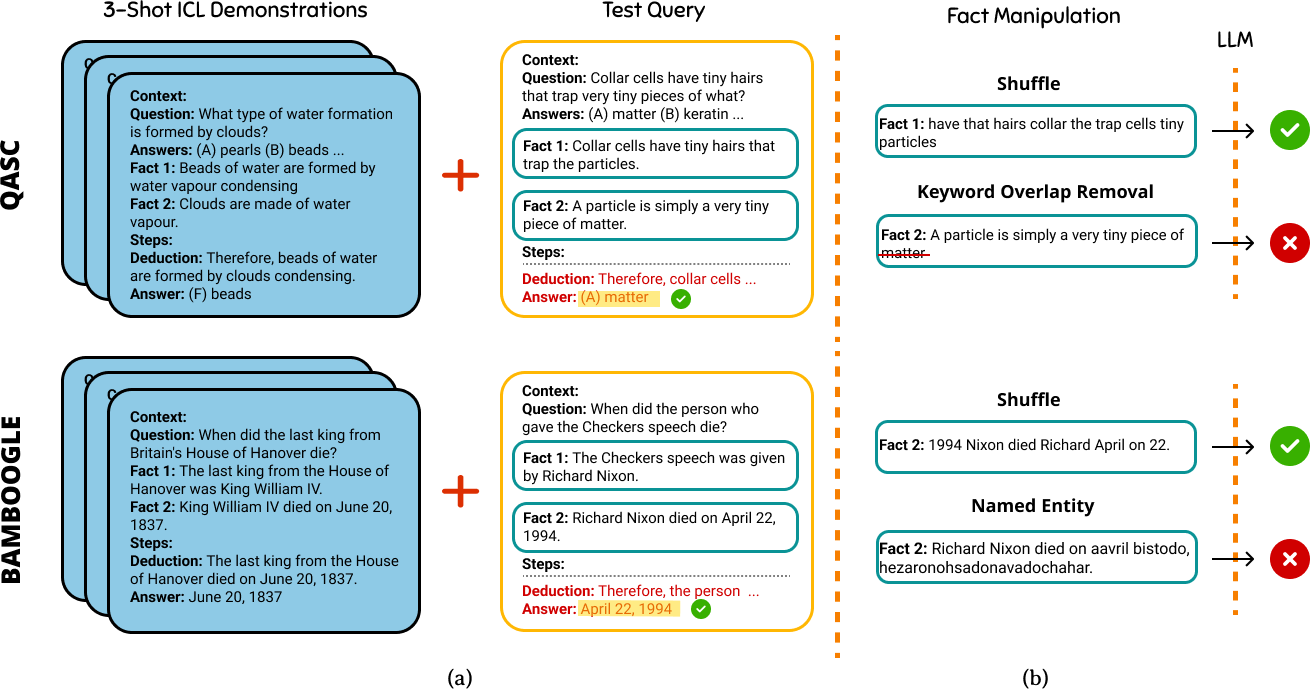}
    \caption{(a) 3-shot In-Context Learning (ICL) prompt for the compositional question answering task. The prompt begins with the instruction \textit{``Follow the demonstrations below to answer the given question''} followed by 3 demonstrations. Each demonstration consists of a ``Context'' with a question, optionally a set of multiple-choice (MC) answers for the QASC dataset \citep{qasc}, two supporting facts (\texttt{fact 1}, \texttt{fact 2}), and a set of ``Steps'' including a ``Deduction'' and the correct answer. The test query contains only the ``Context'' and the LLM needs to generate the ``Steps''. (b) We perform a series of manipulations to either of the facts by shuffling words, removing overlapping keywords, and gibbering Named Entities to control for different sources of exploitation of cues in the input by the models.}
 
    \label{fig:full-prompt}
\end{figure*}
At a high level, \emph{reasoning} refers to the process of an agent deriving information about its environment that extends beyond what is directly observable or retrievable from memory. Large Language Models (LLMs) have shown capabilities of solving complex questions necessitating this very process \citep{llama2, few-shot-learners}. These models can often solve these tasks in few-shot, such as \textit{Chain-of-Thought} (CoT) reasoning \citep{cot, automatic-cot, zhou2023leasttomost}, or in a zero-shot manner \citep{zero_shot_reasoners}. Scaling up LLMs has demonstrated improvements across various multi-step reasoning benchmarks, such as arithmetic, commonsense, and symbolic reasoning \citep{cot, lewkowycz2022solving, emergent_abilities}. Nevertheless, the question of what mechanisms underlie reasoning in these models remains an open one \citep{prystawski2023why, ye-etal-2023-complementary, wang-etal-2023-towards}. Perhaps more pressingly, \textbf{so does the question of whether existing reasoning benchmarks accurately reflect a model's capacity to reason}.

One key aspect of such capabilities is the model’s proficiency in \emph{Transitive Reasoning}. This involves the model’s ability to integrate and logically deduce conclusions from at least two pertinent facts when addressing a specific question (see Figure \ref{fig:full-prompt}). In this paper, we design a set of novel diagnostic experiments using automatic and manual re-annotations of two compositional datasets --QASC \citep{qasc} and Bamboogle 
\citep{press2023measuring}-- to control for the different sources of information the LLMs, namely \llama~\citep{llama2} and \flan~\citep{flant5}, \textbf{might be exploiting in answering compositional questions}. Specifically, our experiments control for (i) named entities in QA pairs; for example, a model looks for dates in the facts when prompted with ``\emph{when}'' in the question, (ii) word/phrase associations or overlaps across sections of the models' input prompt, e.g., removing $B$, in the reasoning chain of $A\rightarrow B$, $B\rightarrow C \Rightarrow A \rightarrow C$, and (iii) the model's exposure to direct answers to the questions during pre-training and/or fine-tuning by using the Bamboogle dataset.

Our initial experiments (Section \ref{sec:experiments}) establish that LLMs perform well with intermediate facts provided (Figure \ref{fig:full-prompt}), demonstrating some transitive reasoning capabilities. Manipulations such as removing overlapping words between facts and questions or shuffling word order within facts show no significant impact on performance. However, removing answer keywords notably decreases model performance, indicating some reliance on these keywords rather than purely relying on transitive reasoning. Experiments controlling for the models’ direct answer knowledge using Bamboogle 
(Section~\ref{sec:bamboogle}) reveal dependency on specific named entities like dates and names for answers. Unlike \llama, \flan~shows more resilience to interference with the named entities of answer tokens, indicating that it engages in a process similar to transitive reasoning due to being knowingly fine-tuned on transitive datasets, though further research is needed to confirm this.

\section{Related Work}

\paragraph{Reasoning} 
LLMs have exhibited certain emergent abilities \citep{emergent_abilities} that 
can be triggered by providing a few demonstrations of CoT manually \citep{cot}, automatically \citep{automatic-cot}, or entirely zero-shot with an instruction, e.g., \textit{`think step-by-step'} \citep{zero_shot_reasoners,flant5}, leading to an increase in performance in downstream tasks that require some form of reasoning. Infusing code in either the pretraining and/or fine-tuning stages has also been shown to help \citep{madaan-etal-2022-language,pal,chen2023program,lyu-etal-2023-faithful}.
Despite their effectiveness in solving reasoning tasks, models usually fail to explore different deductive paths to reach the final answer \citep{saparov2023language}. This can be resolved by oversampling different reasoning paths in generation \citep{wang2023selfconsistency}.

On the models' reasoning analysis, \citet{prystawski2023why} investigate that CoT helps bridge the gap between observations in the pretraining data. \citet{razeghi-etal-2022-impact} finds that models exhibit better numerical reasoning capabilities when the prompt terms are more commonly encountered in the pretraining data. \citet{press2023measuring} and \citet{khot2023decomposed} introduced an iterative prompting method that further improves reasoning beyond CoT. Finally, although increasing the model size usually helps single-hop QA, it does not affect compositional reasoning much \citep{press2023measuring}. 

\paragraph{Reasoning datasets} 
 
    QASC \citep{qasc}, one of the datasets we focus on in this work, is an example of compositional deductive reasoning. It contains science-related multiple-choice questions supported by two statements (facts) that need to be composed to deduce the answer. The answers cannot be directly obtained from a single fact. All the facts follow a simple transitivity rule (Section~\ref{sec:dataset_exp}).
    
    Bamboogle \citep{press2023measuring}, the second dataset of interest, also comprises compositional questions that cannot be answered correctly by a popular search engine. Crucially, it was released after one of the models we experimented with (\mbox{\flan}) was pre-trained. 
   
    StrategyQA \citep{geva-etal-2021-aristotle} also contains multi-step boolean questions which require deducing from two or more facts to answer. However, we exclude it from our benchmarks as preliminary experiments revealed more than 50\% accuracy in answering the questions without the facts, hence making it harder to pinpoint whether the model knows the answers directly, or is required to reason. 

    HotpotQA \citep{yang-etal-2018-hotpotqa} is a multi-hop QA dataset that comprises questions with supporting passages that need to be ingested in a multi-step manner to find the answer. This requires the model
    to perform both extracting of information and reasoning, again possibly hindering identifying a direct link between the question and the answer purely due to reasoning. Finally, GSM8K \citep{gsm8k}, and SVAMP \citep{patel-etal-2021-nlp} are popular mathematical datasets comprising grade school math word problems accompanied by a sequence of deductive steps to solve them. Unlike QASC and Bamboogle, GSM8K and SVAMP do not follow the transitive reasoning style that we aim to study in this paper. Instead, they target the mathematical reasoning of the models. Note that HotpotQA involves finding supporting facts which is not the aim of this paper as we are only interested in the transitive reasoning abilities of the models.

\paragraph{In-Context Learning (ICL)} %In-context learning 
plays an important role in the model's reasoning capabilities \citep{cot}. \citet{min-etal-2022-rethinking} concludes that specifying both the input distribution and the label space in the input prompt is what matters for ICL. \citet{wang-etal-2023-label} show that the labels provided within ICL serve as a reference point for the model during inference. However, \citet{yoo-etal-2022-ground} analyse that the correct input-label mappings could have varying impacts depending on the task at hand. \citet{larger_in-context} show that model size matters in how LLMs deal with ICL: larger models can overwrite their semantic priors if presented with contradictory examples in the input prompt. \citet{webson-pavlick-2022-prompt} find that training a model on corrupted prompts has similar performance to models trained on informative prompts.

\paragraph{Diagnosing Reasoning via Prompting}  
Previous works have also manipulated prompts to uncover reasoning abilities. \citet{ye-etal-2023-complementary} investigate ablating or substituting the input prompt with wrong values. Similarly, \citet{wang-etal-2023-towards} show that incorrect reasoning in the generated CoT steps does not significantly impact model performance; the order of the steps though is crucial. We also study the compositional reasoning behaviour of LLMs in multi-hop questions. We have gone a step further by designing a unique set of experiments aimed at dissecting the model's reliance on linguistic constructs, individual tokens, and their underlying semantics. In contrast to these studies, we are not interested in the effect of ICL: the few-shot demonstrations are kept in their original form. Instead, our emphasis is on modifying the properties of the test queries used to assess our model, allowing us to evaluate its performance under varied conditions without altering the context provided to it. These experiments are designed to discern whether the models engage in compositional deductive reasoning or whether they identify alternative patterns to formulate answers.

\section{Experimental Setup}

\subsection{Task Formulation}
\label{sec:dataset_exp}

We manually inspected and selected datasets that either inherently adhere to the transitive rule of reasoning, such as QASC \citep{qasc}, or can be adjusted with minor re-annotation, like Bamboogle \citep{press-etal-2023-measuring}, to follow: 
\begin{align}
\label{eq:transitivity}
A \rightarrow B, B \rightarrow C \Rightarrow A \rightarrow C    
\end{align}

where $A \rightarrow B$, $B \rightarrow C$ correspond to two facts (henceforth referred to as \texttt{fact 1} and \mbox{\texttt{fact 2}}, respectively), and the deduction is represented by $A \rightarrow C$. This structure mirrors the logical progression inherent in transitive reasoning, with the first two facts serving as premises that lead to the conclusion outlined in the deduction. All prompts can be found in Appendix~\ref{sec:prompts}.

\subsection{Datasets}
\noindent \textbf{QASC} 
features multiple-choice questions (MCQ) answerable through the integration of two facts, leading to a ``Deduction'' (Figure~\ref{fig:full-prompt}). To clarify the derivation from two facts, we prefixed each Deduced Fact with [Therefore,]. Given that each Deduced Fact is logically entailed by the two preceding facts \citep{qasc}, the addition of [Therefore,] at the start serves as a rational and meaningful way to highlight this inferential step. Refer to Appendix~\ref{sec:datasets} for further details.

\noindent \textbf{Bamboogle}

 controls for the models' prior knowledge of the questions and eliminates the biases introduced by an MCQ structured dataset (Section~\ref{sec:bamboogle}). To align with the QASC format, we manually decomposed each question into two related facts by referencing Wikipedia. Questions not found in Wikipedia were omitted, leaving 112 out of the 125 original questions. One of the authors rigorously checked each decomposition to ensure adherence to the transitive rule. For each pair, we then generated a Deduced Fact that maintained the principle of transitivity. Figure~\ref{fig:full-prompt} shows an instance of each of the datasets.

\subsection{Models}

We choose instruction-tuned models that can follow our prompt structure without further fine-tuning. 
In particular, we used \llama~chat (decoder-only architecture; 7B and 13B parameters;  \citealt{llama2}), and 
\flan~ XXL (encoder-decoder; 11B parameters; \citealt{flant5}). \flan~XXL is already fine-tuned on the QASC dataset, allowing us to study whether fine-tuning on a reasoning dataset permits the model to perform some form of transitive reasoning under our diagnostic experiments\footnote{We limited our experiments with \llama~up to 13B parameters, to keep comparisons fair with the largest model from the \flan~family.}. We stick to open-source models for their reproducibility and transparency. For further details refer to Appendix~\ref{sec:implementation}.

\subsection{Metrics}
\label{sec:metrics}
\paragraph{QASC (MCQ) Evaluation} Our evaluation method checks the correctness of the final answer generated by the model. After generating the response, we extract the deductions (if any) and the final answer from the generated response. We use exact matching between the answer choices to calculate accuracy. For instance, if the correct answer is ``(A) matter'' and the model has predicted ``(B) keratin'', we would compare ``(B)'' against ``(A)''.

\paragraph{Bamboogle (non-MCQ) Evaluation} We assess performance on the Bamboogle dataset (Section~\ref{sec:bamboogle} below) using ROUGE-1 \citep{lin-2004-rouge}, since it deviates from the MCQ format. This metric 
evaluates the overlap of unigrams between the gold standard answer and the generated response. 
We refrained from going beyond ROUGE-1 as some models tended to rearrange tokens in certain experiments (for example, generating ``April 30, 1789'' as ``30 April 1789'') or not corresponding with the full answer (generating ``1953'' instead of ``July 27, 1953''); metrics based on n-grams larger than one would fail to take this into account.

\section{QASC and Transitivity}
\label{sec:experiments}
To investigate transitive reasoning (Section~\ref{sec:dataset_exp}) in LLMs, we designed several experiments to analyse their behaviour. Firstly, we explore the performance
when provided with factual information and demonstrations of deduction. Subsequently, we investigate the extent to which knowledge is inherently present within these models, essentially gauging how many answers are pre-existing due to pretraining. We also aim to examine the significance of deductions within these demonstrations. Finally, we inspect the impact of individual facts on the models' ability to deduce the final answer. In all experiments, we used 3-shot ICL \footnote{For QASC we used the dev set to evaluate performance and chose the ICL instances randomly from the training set. For more details on ICL refer to Appendix~\ref{sec:prompts}.}. 

The prompts comprise three sections, beginning with the instruction \textit{``Follow the demonstrations below to answer the given question''}, followed by 3-Shot ICL demonstrations, and ending with the Test Query which prompts the model to generate the response. The overall structure of the prompt is depicted in Figure~\ref{fig:full-prompt}. Depending on the diagnostic experiments, this prompt is modified accordingly (refer to Tables~\ref{tab:prompt-temp-qa}, \ref{tab:prompt-temp-ablation}, and \ref{tab:prompt-temp-shuffle} in Appendix~\ref{sec:prompts}).
Below is the description of prompts for the diagnostic experiments carried out to analyse the models' behaviour dealing with transitivity.
    \paragraph{Full} As illustrated in Figure~\ref{fig:full-prompt}, each demonstration contains a ``Context'' that includes the Question, and a set of multiple-choice (MC) Answers, accompanied by two supporting facts (\texttt{fact 1}, \texttt{fact 2}), and a set of ``Steps'' that crucially comprises a ``Deduction'' before the final Answer. The rationale of the Full prompt is to encourage the model to \textit{deduce} from the facts verbatim before reaching the final answer.
    \paragraph{QA} The demonstrations contain only the Question, the MC Answers, and only the correct Answer as part of the ``Steps''. 
    This prompt aims to check the prior knowledge of the model in answering these questions without any extra information.
    \paragraph{QA (step-by-step)} Similar to \textit{QA}, this prompt contains the \textit{``Think step by step''} Instruction at the beginning, but similarly does not contain any facts in the ``Context'', or a ``Deduction`` step. Inspired by \citet{zero_shot_reasoners} this diagnostic experiment helps identify whether the model does any internal reasoning without explicitly being shown how to do so, e.g., via a ``Deduction'' step.
    \paragraph{QAF} Similar to the \textit{Full} prompt, the model is provided with the facts in the ``Context'' but not the ``Deduction'' step. Therefore, it is tasked with predicting the final answer without generating verbatim any form of reasoning from the facts. This prompt highlights the importance of the deduction step in answering the questions.
    \paragraph{QAF (Fact 1/Fact 2 only)} Identical to the \textit{QAF} prompt, the only difference is that it omits either of the facts. This
    outlines which fact carries more weight for the model's reasoning to reach the final answer. Generally, \texttt{fact 1} is closer to the question, and \texttt{fact 2} contains the answer ($A\rightarrow B$ and $B\rightarrow C$ in Equation~\ref{eq:transitivity}, respectively).

\begin{table}[ht!]
\centering
\small
\begin{tabular}{@{}l@{}c@{~~}c@{~~}c@{}}
\multicolumn{4}{c}{\textbf{QASC Dataset}} \\[5pt]
\midrule
\textbf{Prompt} & \textbf{\llama-13b} & \textbf{\llama-7b} &\textbf{\flan} \\

\midrule
Full & 90 & 74 & 97 \\ \hline
QA & 55 & 43 & 79 \\
QA (step-by-step)& 46 & 37 & 79 \\
QAF & 77 & 56 & 99 \\
QAF (fact 1 only) & 71 & 46 & 94 \\
QAF (fact 2 only) & 60 & 44 & 95 \\

\hline
\end{tabular}
\caption{Accuracy of \llama-13b, \llama-7b, and \flan~XXL on QASC with different diagnostic prompts. \textit{Full} and \textit{QAF} indicate the models' reliance on facts or the deduction step for answering questions. \textit{QA} demonstrates the degree to which the models depend on their inherent knowledge.}
\label{tab:qasc_metrics}
\end{table}

\subsection{Results}
The results from these experiments are depicted in Table~\ref{tab:qasc_metrics}. The first two rows show that \flan~surpasses both \llama-7b and -13b, likely because it has been directly fine-tuned on the QASC dataset. Consistent with the observations made by \citet{emergent_abilities}, the size of the models significantly influences their performance on reasoning tasks.
The \llama~models using the \textit{QA (step-by-step)} prompt perform worse than with \textit{QA}, despite being provided with identical in-context and inference prompts. This could be because the instruction ``Think step by step'' can initiate a different reasoning process more suitable for reasoning tasks other than transitivity. On the other hand, \flan~has been fine-tuned on a series of tasks (including QASC) with the same instruction 
hence, the prompt objective aligns closely with the model's fine-tuning process \citep{flant5, cot}.

Finally, the results with the \textit{QAF} prompt indicate that the \mbox{\llama}~models struggle with reasoning in the absence of deductions within the demonstrations. However, \flan~performs on par with the \textit{Full} prompt, which again could be down to fine-tuning. The last two rows denote that the presence of \texttt{fact 1} is more important in the final answer for the \llama~models but not so much for \flan. Without both facts, executing transitive reasoning becomes unfeasible. This surprising result leads to an intriguing inquiry: \textbf{what information are the models extracting from the facts so that they are
able to outperform the \textit{QA} prompt? }

\begin{table*}[ht!]
\centering
\small
\begin{tabular}{lccc}
\multicolumn{4}{c}{\textbf{QASC Dataset}} \\[5pt]
\midrule
\textbf{Prompt} & \textbf{\llama-13b} & \textbf{\llama-7b} &\textbf{\flan~} \\
\midrule

F1Q Connecting Words Ablation & 85 \color{red}{(-5)} & 45 \color{red}{(-29)} & 93 \color{red}{(-4)}\\
F2Q Connecting Words Ablation & 86 \color{red}{(-4)} & 49 \color{red}{(-25)} & 95 \color{red}{(-2)}\\
F1F2 Connecting Words Ablation& 88 \color{red}{(-2)} & 47 \color{red}{(-27)} & 90 \color{red}{(-7)}\\
F1F2A Keyword Ablation & 75 \color{red}{(-15)}& 40 \color{red}{(-34)} & 83 \color{red}{(-14)} \\
\hline
\end{tabular}
\caption{Accuracy of \llama-13b, \llama-7b, and \flan~XXL on QASC with different ablation prompts. The number in the parentheses represents the delta between the accuracy on the specified prompt and the \textbf{Full} prompt. Models are most reliant on the Answer keywords within the facts to answer the questions.}
\label{tab:ablation_table}
\end{table*}

\section{Does Word Order Matter?}

Previous experiments showed that models benefit significantly when the intermediate facts are provided. This does not mean that the models are engaging in reasoning -- for example, they may be exploiting word overlaps or associations across the questions, facts, and the answers. Reasoning can only proceed from fine-grained, structured meanings of the question, and those of the facts. Therefore, if the models are reasoning over the facts, we would expect them to do significantly worse when the word order in the facts is randomly \textbf{shuffled} (leading essentially to ungrammatical, nonsensical word sequences). We use the following prompt for this experiment:

\paragraph{Shuffled Facts} This prompt follows the \textit{Full} prompt in Section~\ref{sec:experiments}. However, we randomly shuffle every word in \texttt{fact 1} and \texttt{fact 2} delimited by white space (see the first and third instance of Fact Manipulation in Figure~\ref{fig:full-prompt}b).

\begin{figure}[]
    \centering
    \includegraphics[width=\linewidth]{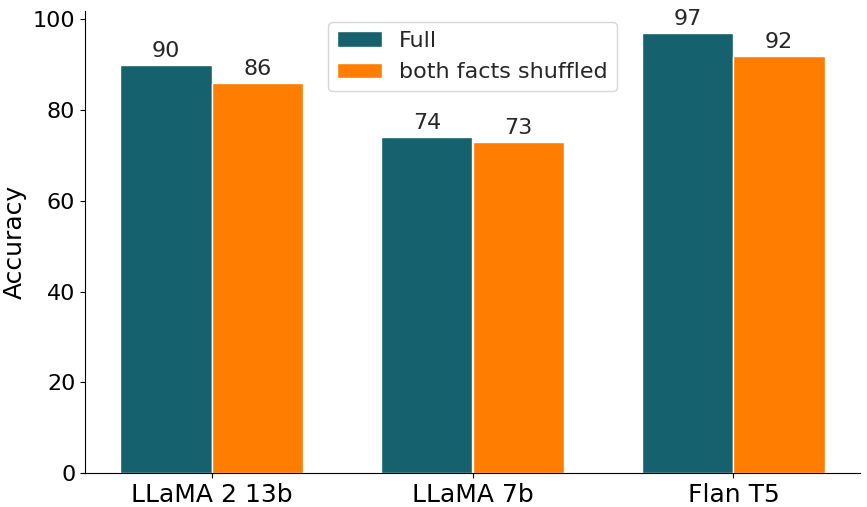}
    \caption{Accuracy of models prompted with the \textit{Shuffled Facts} and \textit{Full} diagnostic prompts. Results show that models are insensitive to word order within facts.} 
    \label{fig:shuffled-facts}
\end{figure}

Figure~\ref{fig:shuffled-facts} shows that shuffling the word order in the facts has minimal impact on the models' performance, one might argue that LLMs are powerful enough to internally restore word order before generating an output. In the following section, we analyse this behaviour in further detail.

\subsection{Can LLMs Restore Word Order Internally?}

To test this, we conducted an experiment, where we prompted our models just to restore the word order of the shuffled facts without performing any question-answering or reasoning task. We hypothesise that if the model is capable of internally restoring the word order of the facts, it should be able to do so when prompted. We begin the experiment with 3-shot ICL demonstrations, where we start with an instruction and provide the shuffled sentence along with the original sentence as the label. The results showed that both models struggled with restoring word order, \llama~and \flan~scoring $10\%$ and $21\%$ respectively. 

By taking a closer look at the results, the models often generated the wrong sequence order, which was not close to the meaning of the original sentence, or generated something that did not have the same words as the original sentence. The ones that the models did restore the word order correctly were the ones that had a short sequence length (e.g. \textit{``a stopwatch is used to measure time''}). This finding confirms that the models are in fact not able to restore the word order of sentences with complex syntactical structures.

Another explanation could be the models' internal knowledge of solving reasoning tasks, making them insensitive to word order. Taking the QA experiment (\ref{tab:qasc_metrics}) into account, helps us understand how much knowledge is inherent within the model. The difference in the model’s performance between the QA and Full experiments indicates that some answers are not inherent in the model, and it relies on external facts to answer these questions. Nevertheless, as shown in Figure~\ref{fig:shuffled-facts} the models were still capable of answering the questions with facts that made no sense, which we showed were not inherent within the model. This intriguing result calls for further investigation into the underlying mechanisms of the models, particularly focusing on how they make transitive deductions. Our next step is to examine whether specific tokens play a pivotal role in the models' ability to reason.

\section{Word/Phrase Associations and Overlaps}

A prominent pattern observed within the QASC dataset is the overlap of words or phrases between the questions and the corresponding facts, as well as among the facts themselves (e.g., the question ``\textit{\textbf{Climate} \textbf{is} \textbf{generally} \textbf{described in terms of} what?}'' and the fact ``\textit{\textbf{Climate is generally described in terms of }temperature and moisture}''). Deliberately omitting connecting words 
between facts effectively disrupts the basis for transitive reasoning, which is designed to impair the model's performance if it depends on this reasoning process. To understand how models depend on these linking tokens, we designed the following experiments that manipulate the \textit{Full} prompt in systematic ways.:

    \paragraph{F1Q Connecting Words Ablation} The mutual words between the Question and \texttt{fact 1} are removed from the latter. As an example, \texttt{fact 1} in Figure~\ref{fig:full-prompt} would be \textit{``the particles''}, but the Question would remain the same.
    
    \paragraph{F2Q Connecting Words Ablation} This prompt is similar to \textit{F1Q Connecting Words Ablation} but with the tokens of \texttt{fact 2} removed.
    
    \paragraph{F1F2 Connecting Words Ablation} All mutual words between \texttt{fact 1} and \texttt{fact 2} are removed.
    \paragraph{F1F2A Keyword Ablation} This prompt is created to analyse the influence of answer choices in the facts on the final generated Answer. In other words, in most cases, the correct answer (choice) is present in one of the facts. In the QASC example from Figure~\ref{fig:full-prompt}, \texttt{fact 2} contains \textit{``matter''} from the choices. As a result, we remove this sequence from \texttt{fact 2} (see the second instance of Fact Manipulation in Figure~\ref{fig:full-prompt}b) to analyse whether the model heavily relies on keywords (this is equivalent to removing $C$ in $B \rightarrow C$; Equation~\ref{eq:transitivity}).

Table~\ref{tab:ablation_table} shows that the smaller \llama~model is more susceptible to the \textit{Connecting Token Ablation}. This suggests that larger models may utilise alternative patterns that enable them to sustain their performance despite the ablation. Additionally, the results suggest that removing all shared tokens from the facts has a minimal effect, even when done in a straightforward manner. This implies that no individual connecting token plays a significant role in influencing model performance. However, the most significant impact on performance is observed when the keyword answer is removed from the facts. This
implies that for certain questions, the models may identify a matching sequence within the answer options and leverage it to generate the answer. Essentially,
the models depend on the presence of answer keywords as a means to simulate the reasoning process. Nevertheless, the accuracy of the models on this dataset can still be attributed to their prior knowledge of the questions.

\section{Models' exposure to direct answers}
\label{sec:bamboogle}

\subsection{Baselines}
To mitigate the impact of models' inherent knowledge of direct answers to questions, we choose the Bamboogle dataset. This dataset consists of questions, which can be decomposed into two questions, the answers to which are provided as facts. To obtain the final answer, the models need to engage in transitive reasoning based on these two facts. Through our initial \textit{QA} experiment, we find that the models have not been exposed to the questions during pre-training or fine-tuning. Moreover, since Bamboogle was released after \flan, it is evident that it has not been fine-tuned on this dataset, ensuring that any performance observed is not the result of prior exposure to the questions. Therefore, it is an ideal dataset to thoroughly examine whether the model is capable of employing transitivity to derive the final answer. The non-multiple choice question (non-MCQ) nature of the dataset further ensures that the model cannot rely on recognising patterns between the choices and the answers to inform its responses. We repeat all the diagnostic prompts on the Bamboogle dataset 
(Table~\ref{tab:Bamboogle_table}).\footnote{We excluded \llama-7b from the results because it mirrors \llama-13b's behaviour; this time we aimed to compare similarly sized models to clarify only the impact of fine-tuning on knowingly relevant reasoning datasets, including e.g., QASC. Note that unlike \flan~we are not aware of the exact dataset \llama~was instruction fine-tuned on.}

\begin{table}[ht!]
\centering
\small
\begin{tabular}{@{}l@{}cc@{}}
\multicolumn{3}{c}{\textbf{Bamboogle Dataset}} \\[5pt]
\midrule
\textbf{Prompt} & \textbf{\llama-13b} &\textbf{\flan~} \\
\midrule
Full & 74 & 96 \\ \hline
QA & 6 & 22\\
QA (step-by-step) & 11 & 6 \\
QAF & 56 & 94\\
QAF (fact 1 only) & 28 & 37 \\
QAF (fact 2 only) & 10 & 95 \\
Full (both facts shuffled) & 63 & 77\\
 F1Q Connecting Words Ablation & 62 & 96 \\
 F2Q Connecting Words Ablation & 71 & 92 \\
 F1F2 Connecting Words Ablation& 70 & 84 \\
\hline
\end{tabular}
\caption{Rouge-1 score of \llama-13b, and \mbox{\flan}~XXL on Bamboogle with different ablation prompts. The Bamboogle dataset controls for the models' prior knowledge to questions. The QA experiment results confirm that both models have not been previously exposed to the questions.}
\label{tab:Bamboogle_table}
\vspace{-3ex}
\end{table}

The low Rouge-1 scores on the \textit{QA} row confirm that the models have not seen much of the dataset, either in part (i.e., the individual facts) or the full question, during pre-training or \mbox{fine-tuning}. The \textit{Full} prompt indicates the models can deduce the correct answers from the facts. The \textit{QAF} prompt also confirms the same findings from the QASC dataset, i.e., \llama-13b needs the deductions within the demonstrations to perform better. When the models are provided with just one of the facts, \flan~demonstrates performance comparable to that achieved with the \textit{Full} prompt when only \texttt{fact 2} is given. Probably 
\texttt{fact 2} invariably contains the answer to the question, in contrast to \texttt{fact 1}, which does not directly provide the answer. Interestingly, \llama-13b is not as good as \flan~in identifying the answer from \texttt{fact 2}.

The results of ablation experiments on \mbox{Bamboogle} closely align with those observed in QASC. However, removing the connecting words between \texttt{fact 2} and \texttt{fact 1} from both facts impairs \flan's performance to a greater extent than was the case in QASC. 

The \textit{Full (both facts shuffled)} results aligned with the observations from QASC dataset, showing that shuffling the tokens within the facts has minimal impact on the final results. Notably, although shuffling disrupts the transitive structure of the facts, the models, particularly \flan~more so than \mbox{\llama-13b}, are still able to find a pattern (distinct from following transitivity) to arrive at the correct answer. Therefore, we search for other patterns the models exploit to sustain performance.

\begin{table}[]
\centering
\small
\begin{tabular}{lcc}
\multicolumn{3}{c}{\textbf{Bamboogle Dataset}} \\[5pt]
\midrule
\textbf{Prompt} & \textbf{\llama-13b} &\textbf{\flan~} \\
\midrule

\multicolumn{1}{@{}l}{\textit{Gibberish}} \\
Full & 49 & 97\\  
Both Facts Shuffled & 10 & 59 \\ \hline
\multicolumn{1}{@{}l}{\textit{Original}} \\
Full & 74  & 96 \\ 
Both Facts Shuffled & 63 & 77\\ \hline
\end{tabular}
\caption{Rouge-1 for \llama-13b and \flan~on the Bamboogle Gibberish dataset with \textit{Full} and \textit{Shuffling} experiments. The second half includes results on the original Bamboogle Dataset for comparison.}
\label{tab:bamboogle_gibberish_table}
\vspace{-3ex}
\end{table}

\subsection{Controlling for Patterns of Named Entities}

To counteract the possibility that models are merely leveraging semantic relationships between questions and answers -- such as seeking out dates when a question starts with \textit{``When''} -- we have lower-cased and shuffled the characters of Proper Names that are answers to questions, and transformed dates into gibberish words within the Bamboogle dataset. The dataset consists exclusively of dates, numbers, or names, all of which have been gibberished (building up $96\%$ of the dataset), with the exception of four responses: one boolean and three nouns. The gibbering targeted numbers, names, and dates specifically to address potential model biases; we name this dataset as ``Bamboogle Gibberish''. We then repeat the \textit{Full} and \textit{Both Facts Shuffled} experiments and compare them with the experiments on the original dataset (Table~\ref{tab:bamboogle_gibberish_table}). 

\mbox{\llama-13b}~ relies on named entities (significant Rouge-1 drop of 25\%), whereas \mbox{\flan}~shows remarkable robustness to our manipulations, thus potentially exhibiting transitive reasoning ability.
This could be down to the fact that fine-tuning helps models generalise to out-of-domain instances
\cite{mosbach-etal-2023-shot}:  explicitly fine-tuning on reasoning datasets --as is definitely the case for \flan-- induces transitive reasoning in the model even with gibberish tokens in the prompt, rather than this behaviour being emergent. 
Adding the shuffling permutation on the Bamboogle gibberish dataset reveals that a portion of the models' ability to identify the correct answer is attributed to the recognition of named entities in the answers. Once these entities were obscured, the models' performance experienced a significant decline across the board (39\% and 38\% for \mbox{\llama-13b} and \flan\footnote{Note that the moderate performance of \flan~(59\%) is probably due to the use of Rouge-1 metric, which is less strict than exact match accuracy. Manual inspection of results paints a worse picture.}, respectively).

\section{Discussion}

Our initial experiments on QASC suggest that LLMs may exhibit a form of reasoning, as shown by strong baseline performances. Specifically, the fact that \flan~has been explicitly fine-tuned on this dataset, explains its performance without demonstrations, hinting at internal reasoning abilities. However, further experiments reveal that these models primarily rely on spotting answer keywords rather than true reasoning. Their correct answers often stem from prior knowledge and the MCQ format of QASC.

We attempt to overcome some of the previous limitation with the use of  Bamboogle.
When evaluated without supporting facts, the models demonstrate limited prior knowledge, as expected. Notably, \flan~
performs well with only \texttt{fact 2}, indicating dependence on specific cues. Like in QASC, shuffling tokens in Bamboogle has minimal impact, suggesting that models may exploit named entities as shortcuts. When controlling for this, \flan~that has knowingly been fine-tuned on relevant reasoning datasets shows capabilities in transitive reasoning, unlike \llama-13b, which relies heavily on named entities. 
Finally, these findings highlight that the ability of models to answer correctly with shuffled word orders largely stems from recognising and using named entities, rather than genuine transitive reasoning.

\section{Conclusion and Future Work}

In this paper, we set out to better understand the underlying processes of LLMs' transitive reasoning through a series of experiments involving the re-annotation of available compositional Question Answering datasets. Experiments revealed that: (a) models not fine-tuned on datasets focused on compositional deductive reasoning perform better when demonstrations include example deductions; (b) there is a noticeable dependence on answer keywords within the facts for question answering, suggesting that performance on reasoning benchmarks should not be taken at face value; and (c) while non-fine-tuned models predominantly rely on named entities to answer questions, models fine-tuned on transitive reasoning tasks demonstrate stronger reasoning capabilities. 

While we identified potential cues that models might exploit when answering transitive questions, we defer a detailed analysis of how specific datasets and task objectives influence transitive reasoning during pre-training and fine-tuning to future research. Given that most models can answer complex questions using few-shot ICL, exploring differences between fine-tuning and ICL regarding reasoning abilities would be interesting future work.

\section{Limitations}
\textbf{Scope:} The scope of the deductive rules that are needed to answer questions in both the QASC and Bamboogle datasets is very limited: all the questions involve the application of \textit{modus ponens} twice in a row; i.e. they exclude all other deductive rules such as \emph{modus tollens}. Any conclusions we draw here are by extension limited in the same way. 

\noindent\textbf{Mechanisms:} This paper does not address the question of \emph{why} LLMs behave as they do. For this, we would need full ablative control over training data and the models themselves. We speculate about the reason behind \flan's robustness to our experimental manipulations; namely that it is because it has been fine-tuned on reasoning datasets. This hypothesis remains to be tested in future work.

\section*{Acknowledgements}
Special thanks to the reviewers, and everyone in the Interaction Lab at Heriot-Watt University, whose invaluable feedback greatly contributed to improving this paper. Ioannis Konstas was supported by the EPSRC project `Equally Safe Online' (EP/W025493/1). Houman Mehrafarin is supported by the Centre for Doctoral Training in Robotics and Autonomous Systems.

\bibliography{anthology_2, custom}
\clearpage 

\appendix

\section{Implementation Details}
\label{sec:implementation}
We used \texttt{transformers} from the HuggingFace library to use the models mentioned in the paper. We also used \texttt{evaluate} from the same library to report the ROUGE-1 scores of the models on the Bamboogle dataset. The accuracy was calculated with the method mentioned in Section~\ref{sec:metrics}. 

We ran our experiments with different seeds and found no inconsistencies in the results. Hence, we ran all experiments with a single seed (a single seed for all potential randomness in the experiments) to control for randomness in the comparisons. We ran our experiments with the following hyper-
parameters: $\texttt{temperature} = 0.7$, $\texttt{top\_p} = 0.75$, $\texttt{top\_k} = 40$, and $\texttt{num\_beams} = 4$. We find that these hyper-parameters are best for our generation task. It is worth noting that we aim to investigate the emerging reasoning ability, rather than to optimise for downstream task performance.

Depending on the model we run our experiments in different batch sizes, but since the experiments are inference only, the batch size does not impact the results. We used batch sizes of 3, 5 and 2 for \llama-13b, -7b, and \flan~respectively. All experiments reported are performed with a single seed, thereby alleviating randomness in comparisons.
\section{Datasets}
\label{sec:datasets}
In this section we provide further details on our chosen datasets.
\paragraph{QASC} The test set of this dataset does not include the supporting facts, which are necessary to our diagnostic experiments, therefore we chose the dev set. We only use the train set to pick our 3-shot ICL demonstrations, and omit the rest. We use the dev set to make sure the models have not seen the questions during training. The total number of samples within this dataset is 926.

\section{Prompts}
\label{sec:prompts}
The first three instances within a dataset were chosen for in-context learning, and they were omitted from the evaluation. Tables~\ref{tab:prompt-temp-qa}, \ref{tab:prompt-temp-ablation}, \ref{tab:prompt-temp-shuffle}, \ref{tab:prompt-temp-gibberish}, and \ref{tab:prompt-restore-word-order} outline the prompt structures used in all experiments.
% ----------------------------PROMPTS----------------------------
\begin{table*}[ht!]
\centering
\renewcommand{\arraystretch}{1.5} % Increase cell height
\resizebox{\textwidth}{!}{
\begin{tabular}{|c|c|l|}
\hline
\textbf{Experiment}
& \multicolumn{1}{c|}{\textbf{Prompt}} & \multicolumn{1}{c|}{\textbf{Details}} \\ \hline
\multirow{5}{*}{\textbf{QA}} & Demonstrations & \makecell[l]{\textbf{Context:}\\[5pt] \textbf{Question:} \texttt{What type of water formation is formed by clouds?} \\ \textbf{Answers:} \texttt{(A) pearls (B) beads ...}\vspace{5pt} \\ \textbf{Steps:}\\ \textbf{Answer:} \texttt{(B) beads} \\[2pt]} \\ \cline{2-3}
& Test & \makecell[l]{\\[-10pt] \textbf{Context:}\\[5pt] \textbf{Question:} \texttt{What is described in terms of temperature} \\ \texttt{and water in the air?} \\ \textbf{Answers:} \texttt{(A) storms (B) climate ...}\vspace{5pt} \\ \textbf{Steps:}\\} \\ \hline
% --------------------------- QAF --------------------------- 
\multirow{6}{*}{\textbf{QAF}} & Demonstrations & \makecell[l]{\\[-10pt]\textbf{Context:}\\[5pt] \textbf{Question:} \texttt{[...]} \\ \textbf{Answers:} \texttt{[...]}\\ \textbf{Fact 1:} \texttt{Beads of water are formed by water vapor condensing.}\\ \textbf{Fact 2:} \texttt{Clouds are made of water vapor.}\vspace{5pt} \\ \textbf{Steps:}\\ \textbf{Answer:} \texttt{(B) beads} \\[2pt]} \\ \cline{2-3}
& Test & \makecell[l]{\\[-10pt]\textbf{Context:}\\[5pt]\textbf{Question:} \texttt{[...]} \\ \textbf{Answers:} \texttt{[...]}\\
\textbf{Fact 1:} \texttt{Climate is generally described in terms of} \\ \texttt{temperature and moisture.}\\ \textbf{Fact 2:} \texttt{Clouds are made of moisture and the moisture} \\
\texttt{is from the water evaporating.}\vspace{5pt} \\ \textbf{Steps:}\\} \\ \hline
% --------------------------- QAF (one fact) --------------------------- 
\multirow{6}{*}{\textbf{QAF (fact 1 only)}} & Demonstrations & \makecell[l]{\\[-10pt]\textbf{Context:}\\[5pt] \textbf{Question:} \texttt{[...]} \\ \textbf{Answers:} \texttt{[...]}\\ \textbf{Fact 1:} \texttt{[...]}\vspace{5pt} \\ \textbf{Steps:}\\ \textbf{Answer:} \texttt{(B) beads} \\[2pt]} \\ \cline{2-3} 
& Test & \makecell[l]{\\[-10pt]\textbf{Context:}\\[5pt]\textbf{Question:} \texttt{[...]} \\ \textbf{Answers:} \texttt{[...]}\\
\textbf{Fact 1:} \texttt{[...]}\vspace{5pt} \\ \textbf{Steps:}\\} \\ \hline
\end{tabular}
}
\caption{Prompts for \textit{QA}, \textit{QAF}, \textit{QAF (Fact 1 only)}, and \textit{QAF (Fact 2 only)} experiments.}
\label{tab:prompt-temp-qa}
\end{table*}
% ---------------------------------- Ablation shuffling ------------------------------

\begin{table*}[ht!]
\centering
\renewcommand{\arraystretch}{1.5} % Increase cell height
\begin{tabular}{|>{\centering\arraybackslash}p{3cm}|c|p{10cm}|}
\hline
\textbf{Experiment}
& \multicolumn{1}{c|}{\textbf{Prompt}} & \multicolumn{1}{c|}{\textbf{Details}} \\ \hline
% --------------------------- F1Q --------------------------- 
\multirow{8}{3cm}{\centering\textbf{F1Q Ablation}} & Demonstrations & \makecell[l]{\\[-10pt]\textbf{Context:}\\[5pt] \textbf{Question:} \texttt{[...]} \\ \textbf{Answers:} \texttt{[...]}\\ \textbf{Fact 1:} \texttt{[...]}\\ \textbf{Fact 2:} \texttt{[...]}\vspace{5pt} \\ \textbf{Steps:}\\ \textbf{Deduction:} \texttt{Therefore, beads of water are}\\ \texttt{formed by clouds condensing.} \\ \textbf{Answer:} \texttt{(B) beads} \\[2pt]} \\ \cline{2-3}
& Test & \makecell[l]{\\[-10pt]\textbf{Context:}\\[5pt]\textbf{Question:} \texttt{What is described in terms of temperature} \\ \texttt{and water in the air?} \\ \textbf{Answers:} \texttt{[...]}\\
\textbf{Fact 1:} \texttt{Climate generally moisture.}\\ \textbf{Fact 2:} \texttt{[...]}\vspace{5pt} \\ \textbf{Steps:}\\} \\ \hline
% --------------------------- F1F2A Keyword Ablation --------------------------- 
\multirow{7}{3cm}{\centering\textbf{F1F2A Keyword Ablation}} & Demonstrations & \makecell[l]{\\[-10pt]\textbf{Context:}\\[5pt] \textbf{Question:} \texttt{[...]} \\ \textbf{Answers:} \texttt{[...]}\\ \textbf{Fact 1:} \texttt{[...]}\\ \textbf{Fact 2:} \texttt{[...]}\vspace{5pt} \\ \textbf{Steps:}\\ \textbf{Deduction:} \texttt{[...]}\\ \textbf{Answer:} \texttt{(B) beads} \\[2pt]} \\ \cline{2-3}
& Test & \makecell[l]{\\[-10pt]\textbf{Context:}\\[5pt]\textbf{Question:} \texttt{[...]} \\ \textbf{Answers:} \texttt{(A) storm (B) climate ...}\\
\textbf{Fact 1:} \texttt{is generally described in terms of} \\ \texttt{temperature and moisture.} \\ \textbf{Fact 2:} \texttt{[...]} \vspace{5pt} \\ \textbf{Steps:}\\} \\ \hline 
\end{tabular}
\caption{Prompts for Keyword ablation}
\label{tab:prompt-temp-ablation}
\end{table*}
% Thanksss
% --------------------SHUFFLE
\begin{table*}[ht!]
\centering
\renewcommand{\arraystretch}{1.5}
\resizebox{\textwidth}{!}{% Increase cell height
\begin{tabular}{|>{\centering\arraybackslash}p{3cm}|c|p{10cm}|}
\hline
\textbf{Experiment}
& \multicolumn{1}{c|}{\textbf{Prompt}} & \multicolumn{1}{c|}{\textbf{Details}} \\ \hline
% --------------------------- Shuffling --------------------------- 
\multirow{7}{3cm}{\centering\textbf{Both Facts Shuffled}} & Demonstrations & \makecell[l]{\\[-10pt]\textbf{Context:}\\[5pt] \textbf{Question:} \texttt{[...]} \\ \textbf{Answers:} \texttt{[...]}\\ \textbf{Fact 1:} \texttt{[...]}\\ \textbf{Fact 2:} \texttt{[...]}\vspace{5pt} \\ \textbf{Steps:}\\ \textbf{Deduction:} \texttt{[...]}\\ \textbf{Answer:} \texttt{(B) beads} \\[2pt]} \\ \cline{2-3}
& Test & \makecell[l]{\\[-10pt]\textbf{Context:}\\[5pt]\textbf{Question:} \texttt{[...]} \\ \textbf{Answers:} \texttt{(A) storm (B) climate ...}\\
\textbf{Fact 1:} \texttt{generally described is temperature in terms of} \\ \texttt{climate moisture and.}\\ \textbf{Fact 2:} \texttt{moisture are made clouds of and the moisture} \\
\texttt{water evaporating is from the.}\vspace{5pt} \\ \textbf{Steps:}\\} \\ \hline
\end{tabular}
}
\caption{Prompts for Shuffled Facts}
\label{tab:prompt-temp-shuffle}
\end{table*}
% --------------------Gibberish
\begin{table*}[ht!]
\centering
\renewcommand{\arraystretch}{1.5} % Increase cell height
\resizebox{\textwidth}{!}{
\begin{tabular}{|>{\centering\arraybackslash}p{3cm}|c|p{10cm}|}
\hline
\textbf{Experiment}
& \multicolumn{1}{c|}{\textbf{Prompt}} & \multicolumn{1}{c|}{\textbf{Details}} \\ \hline
% --------------------------- Gibberish --------------------------- 
\multirow{7}{3cm}{\centering\textbf{Bamboogle Gibberish}} & Demonstrations & \makecell[l]{\\[-10pt]\textbf{Context:}\\[5pt] \textbf{Question:} \texttt{Who was president of the United States in} \\ \texttt{the year that Citibank was founded?} \\ \textbf{Fact 1:} \texttt{Citibank was founded in 1812.}\\ \textbf{Fact 2:} \texttt{The President of the United States in 1812 was} \\\texttt{James Madison.}\vspace{5pt} \\ \textbf{Steps:}\\ \textbf{Deduction:} \texttt{The President of the United States was} \\\texttt{James Madison when Citibank was founded.}\\ \textbf{Answer:} \texttt{James Madison} \\[2pt]} \\ \cline{2-3}
& Test & \makecell[l]{\\[-10pt]\textbf{Context:}\\[5pt]\textbf{Question:} \texttt{Who was the first African American mayor of} \\ \texttt{the most populous city in the United States?} \\ \textbf{Fact 1:} \texttt{The most populous city in the United States is} \\\texttt{New York City.}\\ \textbf{Fact 2:} \texttt{The first African American mayor of} \\\texttt{New York City was ddaiv nkisdni.}\vspace{5pt} \\ \textbf{Steps:}\\} \\ \hline
\end{tabular}
}
\caption{Prompts for Bamboogle Gibberish}
\label{tab:prompt-temp-gibberish}
\end{table*}
% ------------------- WORD ORDER -------------------
\begin{table*}[ht!]
\centering
\renewcommand{\arraystretch}{1.5} % Increase cell height
\begin{tabular}{|>{\centering\arraybackslash}p{3cm}|c|p{10cm}|}
\hline
\textbf{Restore Word Order}
& \multicolumn{1}{c|}{\textbf{Prompt}} & \multicolumn{1}{c|}{\textbf{Details}} \\ \hline

\multirow{7}{3cm}[25pt]{\centering\textbf{Restore Word Order}} & Demonstrations & \makecell[l]{\\[-10pt]\textbf{Context:}\\[5pt] \textbf{Shuffled sentence:} \texttt{of by water formed are water} \\ \texttt{condensing beads vapor} \\ \textbf{Original sentence:} \texttt{beads of water are formed by}\\ \texttt{water vapor condensing}\\[2pt]} \\ \cline{2-3}
& Test & \makecell[l]{\\[-10pt]\textbf{Context:}\\[5pt]\textbf{Shuffled Sentence:} \texttt{varies altitude to climate}\\ \texttt{according} \\ \textbf{Original Sentence:}\\} \\ \hline
\end{tabular}
\caption{Prompt for the Restore Word Order Experiment}
\label{tab:prompt-restore-word-order}
\end{table*}
\end{document}